\documentclass[runningheads]{llncs}
\usepackage{graphicx}
\usepackage{subcaption} 
\usepackage{multirow}
\usepackage{hyperref}
\hypersetup{
    colorlinks=true,
    linkcolor=magenta,
    filecolor=blue,      
    urlcolor=cyan,
    pdftitle={Overleaf Example},
    pdfpagemode=FullScreen,
    }
\usepackage{amsmath} 
\usepackage[misc]{ifsym}

\usepackage{graphicx, hyperref,csquotes,multirow, amsmath, booktabs, bm, amssymb}
\usepackage{comment}
\newcommand{\xv}{{\bm{x}}}
\newcommand{\yv}{{\bm{y}}}
\newcommand{\Tmat}{{\bf T}}
\newcommand{\muv}{{\boldsymbol \mu}}
\newcommand{\nuv}{{\boldsymbol \nu}}

\newcommand{\Pmat}[0]{\ensuremath{{\bf P}} }
\newcommand{\Qmat}[0]{\ensuremath{{\bf Q}} }

\def\rvu{{\mathbf{u}}}
\def\rvv{{\mathbf{v}}}

\begin{document}  
\title{Fair and Accurate Skin Disease Image Classification by Alignment with Clinical Labels}
\titlerunning{PatchAlign: Fair and Accurate Skin Disease Image Classification}

%

\author{
    Aayushman \inst{1}  
    \and
    Hemanth Gaddey\inst{2} 
    \and
    Vidhi Mittal\inst{2} 
    \and
    Manisha Chawla\inst{2} \Letter\\
    \and
    Gagan Raj Gupta\inst{2} 
}


%
\authorrunning{Aayushman et al.}
%
\institute{
Indian Institute of Science Education and Research, Bhopal, India \\
\and
Indian Institute of Technology, Bhilai, India \\
\email{manishachawla96@gmail.com}
}

\maketitle 
%
\begin{abstract}

Deep learning models have achieved great success in automating skin lesion diagnosis. However, the ethnic disparity in these models’ predictions needs to be addressed before deployment of these models. We introduce a novel approach: PatchAlign, to enhance skin condition image classification accuracy and fairness through alignment with clinical text representations of skin conditions. PatchAlign uses Graph Optimal Transport (\texttt{GOT}) Loss as a regularizer to perform cross-domain alignment. The representations thus obtained are robust and generalize well across skin tones, even with limited training samples. To reduce the effect of noise/artifacts in clinical dermatology images, we propose a learnable Masked Graph Optimal Transport for cross-domain alignment that further improves the fairness metrics.

We compare our model to the SOTA model (FairDisCo) on two skin lesion datasets with different skin types: Fitzpatrick17k and Diverse Dermatology Images (DDI). Our proposed approach, PatchAlign, enhances the accuracy of skin condition image classification by 2.8\% (in-domain) and 6.2\% (out-domain) on Fitzpatrick17k  and 4.2\% (in-domain) on DDI compared to FairDisCo. In addition, it consistently improves the fairness of true positive rates across skin tones in all of our experiments.

\textbf{The source code for the implementation is available at the following GitHub repository: \href{https://github.com/aayushmanace/PatchAlign24}{PatchAlign24}, enabling easy reproduction and further experimentation.}

\keywords{Machine Learning Fairness \and Cross-Domain Alignment \and Graph Optimal Transport}

\end{abstract}

\section{Introduction}
\begin{figure}[h]
    \centering
    \begin{subfigure}[b]{0.45\textwidth}
        \centering
        \includegraphics[width=\textwidth,height=0.65\textwidth]{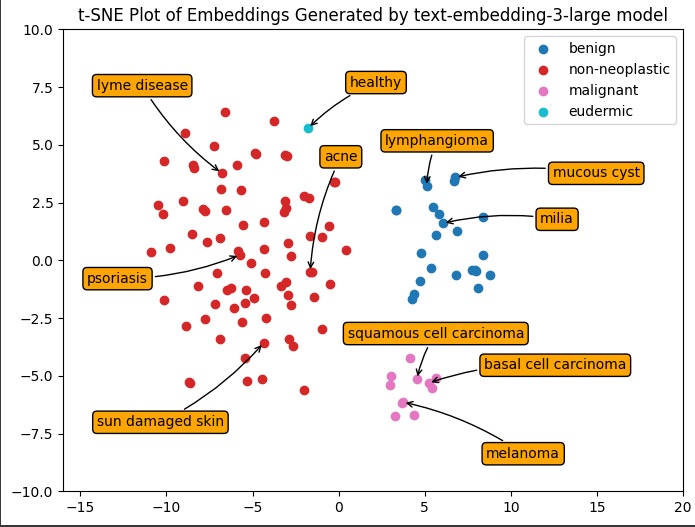}
       \caption{}
        \label{fig:sub1}
    \end{subfigure}
    \hfill
    \begin{subfigure}[b]{0.50\textwidth}
        \centering
        \includegraphics[width=\textwidth,height=0.6\textwidth]{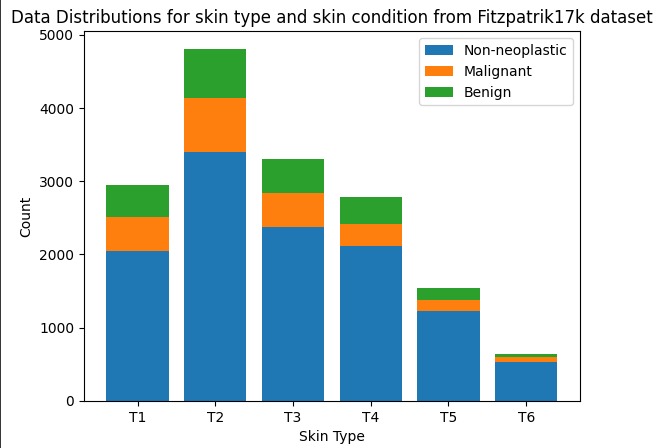}
        \caption{}
        \label{fig:sub2}
    \end{subfigure}
   \caption{(a) t-SNE plot of embeddings of 114 skin conditions in Fitzpatrick17k (b) Data distribution for skin type and skin condition from Fitzpatrik17k dataset. }
    \label{fig:both}
\end{figure}

Skin is among the six most common organs invaded by cancer and early detection and treatment can significantly improve the survival rate of patients~\cite{Beddingfield2003-nb}. Deep learning (DL) offers a promising approach for automated skin disease detection, eliminating the need for manual segmentation and feature extraction ~\cite{kawahara2016deep,esteva2017dermatologist,harangi2018skin,gessert2020skin}. However, recent research~\cite{bhardwaj2021skin,adegun2021deep} highlights the issue of fairness in these models : patients with darker skin tones experience significantly lower diagnosis accuracy compared to those with lighter skin. To address this fairness challenge, researchers have annotated clinical skin images by skin types (tones) and released benchmark datasets like Fitzpatrick17k~\cite{groh2021evaluating} (see Fig.~\ref{fig:both}) and DDI~\cite{daneshjou2021disparities}.

To achieve fairness, DL models should learn representations of skin conditions, that are independent of the skin type. This is difficult because features such as lesion color and visual features that are critical for skin lesion classification can also discriminate an individual’s skin type. Furthermore, clinical skin images often have noise and artifacts (hair, black corners, ruler, etc.)~\cite{yan2023epvt}.

In our proposed approach (Fig.\ref{fig:model}), PatchAlign, a clinical skin image is divided into patches and their representations (embeddings) are aligned with the text embeddings of clinical labels (114 in Fitzpatrick17k and 78 in DDI dataset) (Fig.~\ref{fig:sub1}). We introduce a novel cross-domain alignment technique called \texttt{Masked Graph Optimal Transport (MGOT)}, building on the \texttt{Graph Optimal Transport (GOT)} framework~\cite{GOT}. GOT is designed for cross-domain alignment, excelling in differentiating between various skin conditions and understanding their inter-relationships. MGOT enhances this approach by employing a learnable mask to prioritize disease-relevant patches during alignment. This not only improves accuracy over GOT but also optimizes the GOT loss function. MGOT simultaneously learns to mask irrelevant image patches and update image representations to minimize the alignment cost. Given that most clinical skin images contain areas of healthy skin, incorporating the Eudermic skin label allows MGOT to distinguish between healthy and diseased regions, further enhancing alignment accuracy.
\vskip -5mm

\vskip -5mm
Predictive Quality Disparity (PQD), Equality of Opportunity (EOM), and Demographic Disparity (DPM) are three key fairness metrics~\cite{fairness2021metrics,du2022fairdisco} for evaluation which promote fairness of accuracy, true positive rate, and positive rate prediction respectively across skin types. 

The contributions of our paper are as follows: \\
i. We propose a novel Masked Graph Optimal Transport (MGOT) method as an explicit domain alignment approach for noisy and limited clinical skin images. Our technique incorporates the Eudermic skin label to further enhance alignment accuracy.\\
ii. Extensive experiments, including both in-domain and out-domain evaluations on Fitzpatrick17k and DDI benchmarks.\\ 
iii. Our method, PatchAlign, enhances the accuracy of skin condition image classification by 2.8\% (in-domain) and 6.2\% (out-domain) on Fitzpatrick17k  and 4.2\% (in-domain) on DDI compared to SOTA model FairDisCo. \\
iv. On DDI, it improves PQD by 11.4\%, DPM by 2.8\%, and EOM by 9.2\% and on Fitzpatrick17k improves most metrics compared to FairDisCo. 

\vspace{-2mm}

\section{Related Works}
\vspace{-2mm}
Recent research in skin cancer detection using deep learning \cite{groh2021evaluating} \cite{du2022fairdisco} has highlighted the crucial issue of fairness, with the latter proposing the FairDisCo framework to remove skin type bias from model representations. Additionally, \cite{kalb2023revisiting} reviewed methods for skin tone classification, revealing limitations due to data and methodological inconsistencies. Moreover, \cite{du2020fairness} provided a broader perspective on fairness issues in deep learning, emphasizing interpretability and mitigation strategies. Efforts to address this challenge include automatic skin tone labeling, proposed by \cite{bevan2022detecting}, and fairness-aware model architectures explored by \cite{almuzainit2023toward}. Furthermore, \cite{yan2023epvt} presented a domain generalization method (EPVT) to improve model performance across diverse settings. These related works provide a foundation for our research, which aims to contribute to the development of fair and effective skin cancer detection models.

Furthermore, unlike prior fairness-focused methods that often require extensive data normalisation \cite{Debiasing} or additional human intervention \cite{Yan2023TowardsTS}, PatchAlign leverages a simpler and potentially more effective approach. Our work PatchAlign builds upon the \texttt{GOT}~\cite{GOT} where cross-domain alignment is formulated as a graph matching problem, by representing image patches and skin conditions into a dynamically-constructed graph. The learned model is sparse and interpretable. However, it gives equal importance to all skin patches during alignment. Since clinical skin images have many irrelevant patches, our proposed \texttt{MGOT} decreases their contribution to the cost function.

\vspace{-2mm}
\section{Model Architecture}
\begin{figure}[]
    \centering
    \includegraphics[width=\textwidth,height=0.4\textwidth]{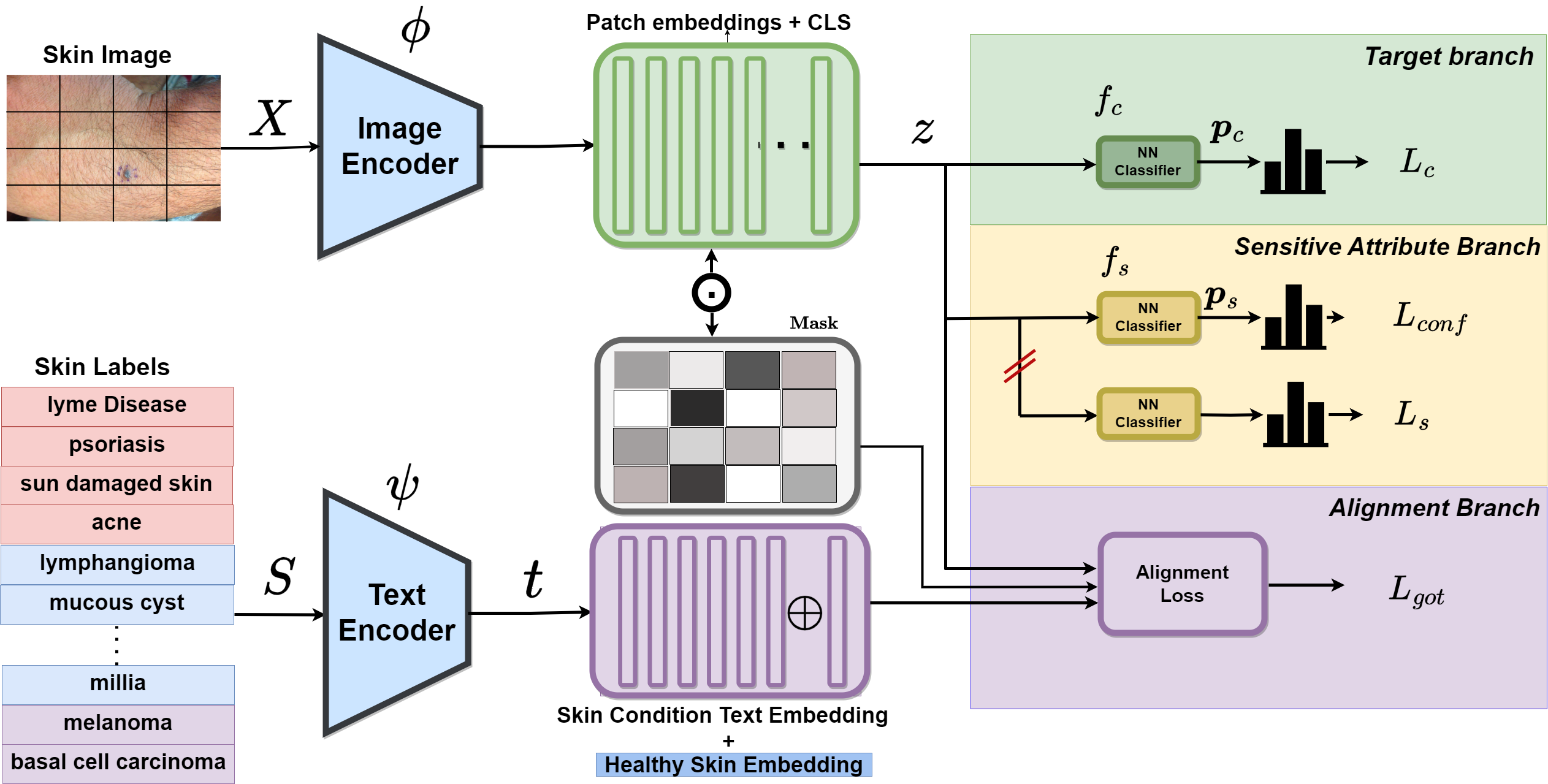}
    \caption{PatchAlign: Our proposed alignment-based skin disease classifier}
    \label{fig:model}
\end{figure}
\vspace{-2mm}
In a multi-class skin lesion classification task with $M$ classes, the model aims to predict a skin condition $y$ based on an RGB skin image $\bm{X} \in \mathbb{R}^{H \times W \times 3}$. We consider the skin type as a sensitive attribute $s \in S$, comprising $N$ groups with diverse types. The objective is to model $p(y|\bm{X})$ while minimizing the influence of $s$. We propose a model architecture (illustrated in Fig.~\ref{fig:model}) comprising an image input, a feature extractor $\phi$, a classifier $f_c$, and a loss function $L_c$. The feature extractor $\phi$ processes the input image to generate a representation $\bm{z} = \phi(\bm{X})$, which is then utilized by the classifier $f_c$ to make predictions regarding the skin condition $\bm{p}_c$. The model is trained using cross-entropy loss $L_c$.

\subsection{Disentangled Representation Learning}
In sensitive attribute branch we utilize disentangled representation learning by using
two different losses. It minimizes a cross entropy loss $L_s$ and confusion loss $ L_{conf} = -\sum_{i=1}^N \frac{1}{N} \log (\boldsymbol{p}_s^i
)$  to effectively remove skin-type information from the learned representations~\cite{du2022fairdisco} so that $f_s$  (in Fig.~\ref{fig:model}) outputs equal probability $\bm{p_s^i}$ for all skin types i. Optimizing $f_s$ with $L_s$ prevents $\boldsymbol{p}_s$ to become a zero vector.

\subsection{Alignment}
For each batch in the data loader, we utilize OpenAI's text-embedding-3-large model~\cite{neelakantan2022text} ($\psi$ in fig:~\ref{fig:model}) to encode all the clinical labels used for alignment. We also added a label for \emph{Eudermic} skin. The alignment is performed between image patches and clinical labels leveraging \texttt{Graph Optimal Transport (GOT)}. Notably, we evaluated other text encoders, including BERT~\cite{bert} and Medical BERT~\cite{medibert}, for clinical label encoding. However,~\cite{neelakantan2022text} achieved superior performance. As a pre-trained model, it requires a one-time computational cost for encoding and has minimal impact on overall model complexity (FLOPs).

We consider two discrete probability distributions $\Pmat$ and $\Qmat \in \mathcal{P}(X)$ (for text embeddings and image embeddings) on space $X \in \mathbb{R}^{d}$. For dynamic construction of graph, we calculate $\{\cos(\xv_i, \xv_j)\}_{i,j}$, then it is thresholded by $\max(\cos(\xv_i, \xv_j)\}_{i,j} \\ - \tau,0)$, where $\tau$ is a threshold hyper-parameter and if the value is positive  an edge is added between node $i$ and $j$. The  \texttt{Graph Optimal Transport (GOT)} distance between $\Pmat$ and $\Qmat$ is formulated as the optimization problem:
\vskip -5mm
\begin{align}
\label{eq:gotd}
\mathcal{D}_{\texttt{\texttt{GOT}}}(\muv,\nuv) = & \min_{\Tmat\in \Pi(\rvu,\rvv)}\sum_{i,i',j,j'} \Tmat_{ij} \Big(\lambda c(\xv_i,\yv_j) \nonumber \\
&+ (1-\lambda) \Tmat_{i'j'} c(\xv_i,\yv_j,\xv_i',\yv_j')\Big) \,.
\end{align}

with $\Tmat \mathbf{1}^m = \rvu$ and $\Tmat^T \mathbf{1}^n = \rvu$, where $\mathbf{1}^m$ is the $m$-dimensional vector of ones. The transport cost $c_{ij} = c(\mathbf{e}_i, \mathbf{l}_j) \geq 0$ between points $\mathbf{e}_i$ and $\mathbf{l}_j$ is defined by a cost function $c(\cdot)$. 

The optimal transport plan $\Tmat$ is trained by minimizing the $\mathcal{D}_{GOT}$ using the iterative Sinkhorn algorithm~\cite{cuturi2013sinkhorn,peyre2017computational} to solve (\ref{eq:gotd}) with an entropic regularizer~\cite{benamou2015iterative} resulting in the iterative Sinkhorn algorithm to reduce the time complexity:
\vskip -3mm
\begin{equation}\label{eq:entropy}
    \min_{\Tmat\in \Pi(\rvu,\rvv)}\sum_{i=1}^n \sum_{j=1}^m \Tmat_{ij} c(\xv_i,\yv_j) + \beta H(\Tmat)\,,
\end{equation}
where $H(\Tmat)=\sum_{i,j}\Tmat_{ij}\log \Tmat_{ij}$, and $\beta$ is the hyper-parameter controlling the importance of the entropy term. 

\textbf{Masked \texttt{GOT}}: It is the same as in eq: [\ref{eq:gotd}, \ref{eq:entropy}] with ( \(\Tmat \text{ replaced by }  M \odot \Tmat\) ) for both objective function and constraints. We generate the Mask weights from patch embeddings using a generator network with sigmoid activation function so that $M_{ij} \in [0,1]$. The generated weights from the mask are then multiplied to patch embeddings. Thus, patches that align better with skin condition representations have higher weights than noisy patches.

Thus the final Loss function is:
\vskip -6mm
\begin{equation}\label{equation: total}
    L_{total}=L_c(\theta_\phi,\theta_{f_c})+\alpha L_{conf}(\theta_\phi,\theta_{f_s})+L_s(\theta_{f_s})+\beta L_{\texttt{GOT}}(\theta_\phi,\theta_\psi, \texttt{Mask}).
\end{equation}
We use $\alpha$ and $\beta$ to adjust contributions of confusion loss and alignment loss. Notice $L_s$ is only used to optimize $f_s$.

\subsection{Multi-Task Approach for learning better representations}
\label{section:multitask}
In addition to PatchAlign, we investigate multi-task learning (MTL)~\cite{multitasking} to learn high-quality representations for skin cancer detection. MTL involves training a model on multiple tasks simultaneously. In our case, the two tasks are: 1) predicting meta-labels and 2) predicting the skin condition.

We modify the GOT loss function from Eq. \ref{equation: total} for MTL. The GOT loss is replaced with a combination of two losses: a predictive cross-entropy loss for the skin conditions and a contrastive loss for the skin labels used for classification, similar to FairDisCo~\cite{du2022fairdisco}. This approach implicitly encourages alignment between the tasks by leveraging shared representations. Here, MTL implicitly encourages alignment, whereas PatchAlign explicitly addresses alignment through the GOT loss, potentially leading to more robust model performance.

\section{Implementation}
\vskip -1mm
\subsection{Dataset Explanation}
We study two skin lesion datasets: Fitzpatrick17k dataset~\cite{groh2021evaluating} and Diverse Dermatology Images (DDI) dataset~\cite{daneshjou2021disparities}. \cite{groh2021evaluating} contains 16,577 clinical images annotated with their Fitzpatrick17k skin-type labels which is a six-point scale initially developed for classifying sun reactivity of skin and adjusting clinical treatment according to skin phenotype~\cite{fitzpatrick1988validity}. There are 114 different skin conditions(clinical labels), and each one has at least 53 images. They further grouped these skin conditions into 3 categories: malignant, non-neoplastic, benign. See Fig.~\ref{fig:both} for the data distribution.
The DDI dataset contains 656 images with diverse skin types and pathologically confirmed skin condition labels, including 78 detailed disease labels and malignant identification. They grouped 6 Fitzpatrick scales into 3 groups: Fitzpatrick-12, Fitzpatrick-34, and Fitzpatrick-56, where each contains a pair of skin-type classes, i.e., \{1,2\}, \{3,4\} and \{5,6\}, respectively.

\subsection{Metrics}
\label{section:metrics}
\begin{enumerate}
  
    \item Predictive Quality Disparity (PQD)~\cite{fairness2021metrics} is computed as the ratio between the lowest accuracy to the highest accuracy across each sensitive group j in S (where S is a set of skin type)  
    \vskip -3mm
\begin{equation}
    PQD = \frac{\min(acc_j, j \in S)}{\max(acc_j, j \in S)} \;\;    
\end{equation}

    \item Demographic Disparity (DP)~\cite{fairness2021metrics} computes the percentage diversities of positive outcomes for each sensitive group. We compute DPM across multiple skin conditions, \(m \in \{1, 2, \dots , M\}\), as follows:
    \vskip -3mm
\begin{equation}
    DPM = \frac{1}{M}\sum^{M}_{i=1}\frac{\min [p(\hat {y}=i|s=j), j \in S]}{\max [p(\hat {y}=i|s=j), j \in S]}
\end{equation}

    \item Equality of Opportunity (EO)~\cite{fairness2021metrics} asserts that different sensitive groups should have
similar true positive rates. We compute EOM across multiple skin conditions as follows:
\vskip -3mm
\begin{equation}
    EOM = \frac {1}{M}\sum _{i=1}^{M}{\frac {\min [p(\hat {y}=i|y=i,s=j), j \in S]}{\max [p(\hat {y}=i|y=i,s=j), j \in S]}}
\end{equation}

\end{enumerate}
\vskip -3mm
where y is the ground-truth skin condition label and \(\hat{y}\) is the model prediction. A model is fairer if it has higher values for the above three metrics. EOM, ensuring a consistent true positive rate across diverse skin types, stands out as a crucial fairness metric. In contrast, DPM focuses solely on achieving a comparable positive prediction rate, not necessarily true positive rate. Since we do not want to increase false positives, EOM is a more critical metric than DPM.



\subsection{Model Training}
For the Fitzpatrick17k dataset, we carry out a three-class classification and follow Groh et al.~\cite{groh2021evaluating} in performing two experimental tasks: In-domain and Out-domain classification. The first is an in-domain classification, in which the train and test sets are randomly split in an 8:2 ratio (13261:3318).
For each batch in the data loader, we utilize text-embedding-3-large model to generate embeddings of all clinical labels. In addition, we add the label \emph{eudermic} skin because many of the patches have healthy skin. Source code is available at \url{https://github.com/aayushmanace/PatchAlign24}

For the DDI dataset, we perform an in-domain binary classification (malignant vs non-malignant) using the same train-test ratio of 8:2 (524:132). For all the models, we use a pre-trained ViT.
Images are augmented through random cropping, rotation, and flipping to boost data diversity, then resized to 224 × 224 × 3. We use Adam~\cite{kingma2014adam} optimizer to train the model with an initial learning rate $1 \times 10^{-4}$, which changes through a linear decay scheduler whose step size is 2, and decay factor $\gamma$ is 0.8. We deploy models on a single NVIDIA RTX A6000 and train them with a batch size of 32. We set the training epochs for the Fitzpatrick17k dataset to 20 and the DDI dataset to 20.

\section{Results}
\subsection{Overall Performance}
For all the experiments, we report individual accuracies for all skin types, their average, and fairness metrics, as explained in Section \ref{section:metrics} on DDI and Fitzpatrick17k Dataset. We employed several strong baseline models to ensure a comprehensive evaluation. We began with a simple pre-trained ResNet-18 architecture(BASE) as a foundation. The BASE model uses pre-trained ResNet-18 as the feature extractor and only cross-entropy loss on the final labels for optimization. We then incorporated the results from FairDisCo~\cite{du2022fairdisco}, a well-established approach, as a baseline for comparison. Finally, to demonstrate our strategy's effectiveness, we included a multi-tasking model's performance(MTL) as an additional baseline.
\vskip -5mm
\begin{table}[]
\centering
\caption{In-Domain classification on DDI Dataset}
\setlength{\tabcolsep}{1pt} 
\renewcommand{\arraystretch}{1.0} 
\resizebox{\textwidth}{!}{ 
    \begin{tabular}{|l|c|c|c|c|c|c|c|}
    \hline
    \multirow{2}{*}{Model} & \multicolumn{4}{c|}{Accuracy (\%) $\pm$ Std-dev.} & \multicolumn{3}{c|}{Fairness Metrics} \\
    \cline{2-5} \cline{6-8}
    & Avg & T12 & T34 & T56 & $PQD$ & $DPM$ & $EOM$ \\
    \hline
    BASE & 82.4 $\pm$ 1.5 & 83.3 $\pm$ 1.0 & 74.6 $\pm$ 5.7 & 89.7 $\pm$ 2.2 & 77.0 $\pm$ 1.9 & 75.2 $\pm$ 13.3 & 58.7 $\pm$ 4.3  \\
    FairDisCo & 83.8 $\pm$ 0.4 & 88.6 $\pm$ 0.1 & 71.7 $\pm$ 2.2 & 92.0 $\pm$ 2.8 & 78.0 $\pm$ 4.5 & 72.8 $\pm$ 12 & 63.7 $\pm$ 3.5  \\
    MTL & 82.3 $\pm$ 0.4 & 79.4 $\pm$ 3.3 & 82.6 $\pm$ 3.1 & 85.6 $\pm$ 1.5 & 91.4 $\pm$ 2.7 & 57.7 $\pm$ 5.9 & 77.2 $\pm$ 4.6 \\
    PatchAlign & 87.4 $\pm$ 1.2 & 89.6 $\pm$ 2.6 & 80.3 $\pm$ 5.7 & 92.3 $\pm$ 1.3 & 86.9 $\pm$ 6.1 & 74.9 $\pm$ 12 & 69.6 $\pm$ 1.7   \\
    \hline
    \end{tabular}
}
\label{table:ddi_table}
\end{table}

 \textbf{DDI Dataset:} From Table \ref{table:ddi_table} we can see that PatchAlign consistently improves all metrics over FairDisCo.  It improves average accuracy over FairDisCo by 4.2\%, PQD by 11.4\%, DPM by 2.8\%, and EOM by 9.2\%. Notably, since DDI contains a limited number of samples per skin condition, we conclude that PatchAlign can learn high-quality representations that are independent of the skin type. 

\textbf{Fitzpatrick17k Dataset:} Table \ref{table:main_table} presents a comprehensive comparison
of various baseline and other existing fairness methods RESM \cite{rewt_resm}, REWT\cite{rewt_resm}and ATRB\cite{attribute_aware} on Fitzpatrik17k dataset. RESM (Resampling Algorithm) organizes samples with similar skin types and conditions into unified groups and then oversamples minorities and undersamples majorities to construct a balanced dataset. REWT (Reweighting Algorithm) aims to make skin types and conditions independent of each other. ATRB (Attribute-aware Algorithm) adds skin type information to the model so that its prediction will not be dominated by major groups. From Table \ref{table:main_table},  we see a significant increase of 4.2\% in overall accuracy and fairness metrics, especially DPM by 8.4\% and EOM by 14.7\% using PatchAlign. Moreover, we achieved the highest average test accuracy using MTL. We also note that our methods perform significantly better even on Darker skin tones that have a much smaller number of samples in the training set. Notably, PatchAlign achieves the highest accuracy (91.4\%) on skin type T6, increasing the accuracy score from all other previously proposed models by more than 7\%.
\vskip -5mm
\subsection{\textbf{Ablation Study}}
We conduct ablation studies to analyze various components of our different models, as detailed in the last four rows of Table \ref{table:main_table}. For a fair comparison, all models utilize the pre-trained ViT~\cite{dosovitskiy2021image} as the image encoder. We begin with the Baseline Model, which employs only the cross-entropy loss (BASE ViT). Subsequently, we integrate GoT with the BASE ViT, and finally, we experiment with FairDisco using the ViT encoder. Furthermore, we have also compared with {\(\text{PatchAlign}^{\oslash}\)}, which does not use \emph{eudermic} skin type during cross-domain alignment. Upon introducing alignment between labels and patches, PatchAlign increases the average accuracy score along with the fairness metrics. We also observe that the addition of the \emph{eudermic} label results in the highest EOM Metric. Some patches of the image that have healthy skin align with the \emph{eudermic} label and not with the labels of other diseases, as was happening earlier. Thus, false positives decrease, in turn increasing true positives and the EO Metric. We have also done an ablation study for entropy value $\beta$ and found 0.8 to yield maximum accuracy.

\vskip -2mm
\begin{table}[t]
\captionsetup{justification=centering} 
\caption{In-Domain classification on Fitzpatrick17k Dataset}
\centering
\setlength{\tabcolsep}{3pt} 
\renewcommand{\arraystretch}{1.0} 
\begin{tabular*}{\linewidth}{|l|c|c|c|c|c|c|c|c|c|c|}
\hline
\multirow{2}{*}{Model} & \multicolumn{7}{c|}{Accuracy (\%) $\uparrow$  } & \multicolumn{3}{c|}{Fairness Metrics (\%)} \\
\cline{2-8} \cline{9-11}
& Avg & T1 & T2 & T3 & T4 & T5 & T6 & PQD & DPM & EOM \\
\hline
BASE(ResNet-18) & 85.0  & 82.6 & 81.7 & 85.0 & 88.8 & 90.7 & 82.8 & 88.9 & 52.5 & 64.0\\
RESM & 85.1 & 82.5 & 81.8 & 86.2 & 89.0 & 91.2 & 82.3 & 89.0 & 50.7 & 62.0\\
REWT & 85.6 & 83.4 & 83.0 & 86.2 & 89.0 & 90.6 & 83.5 & 90.8 & 50.0 & 62.9\\
ATRB & 84.9 & 82.4 & 82.0 & 85.4 & 88.9 & 90.4 & 84.0 & 90.0 & 46.7 & 60.4\\
FairDisCo & 85.1 & 82.2 & 82.1 & 86.2 & 89.4 & 90.0 & 83.2 & 90.2 & 51.2 & 63.8 \\
MTL & \underline{88.3}&	\textbf{86.0}&	\textbf{86.3}&	88.9&\textbf{91.5} & 91.3 & \underline{90.0} &	\textbf{93.7}&	50.5&	71.2 \\ 
PatchAlign & \textbf{88.6} & 82.3 & \underline{85.6} & \underline{89.2}& \underline{91.4}& \textbf{91.3}& \textbf{91.4} & 90.0 & \underline{55.5} & \textbf{74.8}\\
\hline
BASE(ViT) & 86.0 &83.5 & 82.9 & 86.9 & 90.5 & 90.5 & 84.6 & 89.4 &47.0 &59.2 \\
BASE(ViT)+GOT & 86.6 &81.8 &84.2 &88.4& 91.4 &91.3 &83.3& 89.4 &47.0 &59.2 \\
FairDisCo$^{\#}$& 87.0 & 83.2 & 84.7 & 89.1 & 90.8 & 92.6 & 88.4 & 89.8 & 48.3 & 68.1 \\
PatchAlign$^{\oslash}$ & 87.4 & \underline{84}.1& 84.7& \textbf{90.3}& 90.7 & \underline{90.0} &84.2 & \underline{92.7} &\textbf{62.7}& \underline{72.6} \\ 

\hline
\end{tabular*}
\label{table:main_table}
\end{table}
\vskip -2mm

\begin{table}[]
\centering
\caption{Out-Domain Classification on the Fitzpatrick17k Dataset}
\small
\setlength{\tabcolsep}{4pt} 
\begin{tabular}{|c|c|ccccccc|ccc|}
\hline
\multirow{2}{*}{E} & \multirow{2}{*}{Model} & \multicolumn{7}{c|}{Accuracy} & \multicolumn{3}{c|}{Fairness Metric} \\ \cline{3-12} 
 &  & Avg & T1 & T2 & T3 & T4 & T5 & T6 & PQD & DPM & EOM \\
\hline
\multirow{2}{*}{A} & FairDisco & \multicolumn{1}{c|}{79.5} & \multicolumn{1}{c|}{-} & \multicolumn{1}{c|}{-} & \multicolumn{1}{c|}{80.2} & \multicolumn{1}{c|}{79.4} & \multicolumn{1}{c|}{78.1} & 79.0 & \multicolumn{1}{c|}{96.9} & \multicolumn{1}{c|}{\textbf{73.5}} & \textbf{71.2}  \\ \cline{2-12} 
 & \textbf{PatchAlign} & \multicolumn{1}{c|}{\textbf{84.0}} & \multicolumn{1}{c|}{\textbf{-}} & \multicolumn{1}{c|}{\textbf{-}} & \multicolumn{1}{c|}{\textbf{84.2}} & \multicolumn{1}{c|}{\textbf{84.0}} & \multicolumn{1}{c|}{\textbf{83.1}} & \textbf{84.6} & \multicolumn{1}{c|}{\textbf{98.3}} & \multicolumn{1}{c|}{{61.7}} & 68.2 \\ \hline
\multirow{2}{*}{B} & FairDisco & \multicolumn{1}{c|}{78.3} & \multicolumn{1}{c|}{73.4} & \multicolumn{1}{c|}{ 77.9} & \multicolumn{1}{c|}{-} & \multicolumn{1}{c|}{-} & \multicolumn{1}{c|}{86.7 } & 83.7 & \multicolumn{1}{c|}{84.6} & \multicolumn{1}{c|}{53.7} & 64.6 \\ \cline{2-12} 
 & \textbf{PatchAlign} & \multicolumn{1}{c|}{\textbf{82.4}} & \multicolumn{1}{c|}{\textbf{78.6}} & \multicolumn{1}{c|}{\textbf{82.1}} & \multicolumn{1}{c|}{\textbf{-}} & \multicolumn{1}{c|}{\textbf{-}} & \multicolumn{1}{c|}{\textbf{89.5}} & \textbf{86.1} & \multicolumn{1}{c|}{\textbf{87.8}} & \multicolumn{1}{c|}{\textbf{70.7}} & \textbf{75.1} \\ \hline
\multirow{2}{*}{C} & FairDisco & \multicolumn{1}{c|}{71.5} & \multicolumn{1}{c|}{65.3} & \multicolumn{1}{c|}{68.9} & \multicolumn{1}{c|}{73.1} & \multicolumn{1}{c|}{80.4} & \multicolumn{1}{c|}{-} & - & \multicolumn{1}{c|}{81.2} & \multicolumn{1}{c|}{59.3} & 77.5 \\ \cline{2-12} 
 & \textbf{PatchAlign} & \multicolumn{1}{c|}{\textbf{77.6}} & \multicolumn{1}{c|}{\textbf{74.4}} & \multicolumn{1}{c|}{\textbf{75.6}} & \multicolumn{1}{c|}{\textbf{78.2}} & \multicolumn{1}{c|}{\textbf{83.5}} & \multicolumn{1}{c|}{-} & - & \multicolumn{1}{c|}{\textbf{89.1}} & \multicolumn{1}{c|}{\textbf{74.7}} & \textbf{78.1} \\ \hline
\end{tabular}

\label{table:out_domain1}
\end{table}

\subsection{Out Domain Classification on Fitzpatrick17k Dataset}
We perform an out-domain classification where we train the model on all the images of two skin types and test on other skin types. From Table \ref{table:out_domain1}, we can see PatchAlign performs significantly better, reporting a significant increase in all three cases. Our proposed approach, PatchAlign, enhances
the accuracy of skin condition image classification by 6.2\% in out-domain on Fitzpatrick17k compared to FairDisCo.

\vspace{-2mm}
\section{Conclusion}
In this paper, we have proposed a novel Masked Graph Optimal Transport (MGOT) as an explicit domain alignment approach for noisy and limited clinical skin images. Our method, PatchAlign, enhances the accuracy and fairness of skin condition image classification on both Fitzpatrick17k and DDI datasets compared to SOTA model FairDisCo. Significantly, PatchAlign performed well even with less data. Its performance on under-represented T56 images indicates its effectiveness in capturing and learning from diverse data distributions, which is crucial for real-world applications where such variations are common.

    \subsubsection*{\textbf{Disclosure of Interests.}} 
    \small The authors have no competing interests to declare that are relevant to the content of this article.

\bibliographystyle{plain}
\bibliography{Paper-2609}
\end{document}